# "Ideal Parent" Structure Learning for Continuous Variable Networks

**Iftach Nachman**[1]    **Gal Elidan**[1]    **Nir Friedman**
School of Computer Science & Engineering, Hebrew University
{*iftach,galel,nir*}@*cs.huji.ac.il*

## Abstract

In recent years, there is a growing interest in learning Bayesian networks with continuous variables. Learning the structure of such networks is a computationally expensive procedure, which limits most applications to parameter learning. This problem is even more acute when learning networks with hidden variables. We present a general method for significantly speeding the structure search algorithm for continuous variable networks with common parametric distributions. Importantly, our method facilitates the addition of new hidden variables into the network structure efficiently. We demonstrate the method on several data sets, both for learning structure on fully observable data, and for introducing new hidden variables during structure search.

## 1  Introduction

Bayesian networks in general, and continuous variable networks in particular, are being used in a wide range of applications, including fault detection (*e.g.*, [12]), modeling of biological systems (*e.g.*, [6]) and medical diagnosis (*e.g.*, [18]). A key appeal of probabilistic models is that they can be learned from data. This allows us to overcome lack of expert knowledge about domains, adapt models to changing environment, and also lead to scientific discoveries.

A key task in learning is adapting the structure of the network based on observations. This is typically treated as a combinatorial optimization problem, which is addressed by heuristic search procedures, such as greedy hill climbing. This procedure examines local modifications to single edges at each step, evaluates them, and proceeds to apply the one that leads to the largest improvement. This is repeated until reaching a local maximum. Yet, even with this simple greedy approach, structure learning remains a computational challenge in many real-life domains due to the large number of possible modifications that can be evaluated, and the cost of evaluating each one. When learning in the presence of missing values, the problem is even more complicated, as it requires non-linear optimization to evaluate different candidates during the search. This is especially problematic when we want to add new hidden variables during the learning process.

In here we focus on learning continuous variable networks, which are crucial for a wide range of real-life applications. When we learn *linear Gaussian* networks [9], we can use sufficient statistics to summarize the data, and a closed form equation to evaluate the score of candidate families. In general, however, we are also interested in non-linear interactions. These do not have sufficient statistics, and require applying parameter optimization to evaluate the score of a candidate family. These difficulties severely limit the applicability of standard heuristic structure search procedures to rich non-linear models.

We present a general method for speeding search algorithms for structure learning in continuous variable networks. Our method can be applied to many forms of a unimodal parametric conditional distribution, including the linear Gaussian model as well as many non-linear models. The ideas are inspired from the notion of *residues* in regression [14], and involve the notion of "ideal parents". For each variable, we construct an *ideal parent profile* of a new hypothetical parent that would lead to the best possible prediction of the variable. We then use this profile to efficiently select potential candidate parents that have a similar profile of values. Using basic principles, we derive a similarity measure that can be computed efficiently and that approximates the improvement in score that would result from the addition of a candidate hidden parent. This provides us with a fast method for scanning many potential parents and focus more careful evaluation (scoring) to a smaller number of promising candidates.

The ideal parent profiles we construct during search also provide new leverage on the problem of introducing new hidden variables during structure learning. Basically, if the ideal parent profiles of several variables are sufficiently similar, and are not similar to the profile of an existing variable in the current model, we can consider adding a new hidden variable that serves as a parent of all these variables. The ideal profile allows us to estimate the impact this new variable will have on the score, and suggest the values it takes in each instance. And so, the method provides a guided approach for introducing new variables during search and allows to contrast them with alternative search steps in a computationally efficient manner.

---

[1]These authors contributed equally to this manuscript

## 2　Continuous Variable Networks

Consider a finite set $\mathcal{X} = \{X_1, \ldots, X_n\}$ of random variables. A *Bayesian network* (BN) is an annotated directed acyclic graph $G$ that represents a joint distribution over $\mathcal{X}$. The nodes correspond to the random variables and are annotated with a conditional probability density (CPD) of the random variable given its parents $\mathbf{U_i}$ in the graph $G$. The joint distribution is the product over families

$$P(X_1, \ldots, X_n) = \prod_{i=1}^{n} P(X_i | \mathbf{U_i})$$

Unlike the case of discrete variables, when the variable $X$ and some or all of its parents are real valued, there is no representation that can capture all conditional densities. A common choice is the use of *linear Gaussian* conditional densities [9], where each variable is a linear function of its parents with Gaussian noise. When all the variables in a network have linear Gaussian conditional densities, the joint density over $\mathcal{X}$ is a multivariate Gaussian [11]. In many real world domains, such as in neural or gene regulation network models, the dependencies are known to be non-linear (for example, a saturation effect is expected). In these cases we can still use Gaussian conditional densities, but now the mean of the density is expressed as a non-linear function of the parents (for example, a sigmoid).

Given a training data set $\mathcal{D} = \{\mathbf{x}[1], \ldots, \mathbf{x}[M]\}$, where the $m$'th instance $\mathbf{x}[m]$ assigns values to the variables in $\mathcal{X}$, the problem of learning a Bayesian network is to find a structure and parameters that maximize the likelihood of $\mathcal{D}$ given the graph, typically along with some regularization constraints. Given $\mathcal{D}$ and a network structure $G$, we define

$$\ell(\mathcal{D} : G, \theta) = \log P(\mathcal{D} : G, \theta) = \sum_m \log P(\mathbf{x}[m] : G, \theta)$$

to be the log-likelihood function, where $\theta$ are the model parameters. In estimating the *maximum likelihood* parameters of the network, we aim to find $\hat{\theta}$ that maximize this likelihood function. When the data is *complete* (all variables are observed in each instance), the log-likelihood can be rewritten as a sum of *local* likelihood functions,

$$\ell(\mathcal{D} : G, \theta) = \sum \ell_i(\mathcal{D} : \mathbf{U}_i, \theta_i)$$

where $\ell_i(\mathcal{D} : \mathbf{U}_i, \theta_i)$ is a function of the choice of $\mathbf{U}_i$ (the set of parents of $X_i$ in $G$), and the parameters $\theta_i$ of the corresponding CPD: it is the log-likelihood of regressing $X_i$ on $U_i$ in the data set with the particular choice of CPD. Due to this decomposition, we can find the maximum likelihood parameters of each CPD independently by maximizing the local log-likelihood function. For some CPDs, such as linear Gaussian ones, there is a closed form expression for the maximum likelihood parameters. In other cases, finding these parameters is a continuous optimization problem that is typically addressed by gradient methods.

Learning the structure of a network is a significantly harder task. The common approach is to introduce a scoring function that balances the likelihood of the model and its complexity and then attempt to maximize this score using a heuristic search procedure that considers local changes (*e.g.*, adding and removing edges). A commonly used score is the *Bayesian Information Criterion* (BIC) score [17]

$$BIC(\mathcal{D}, G) = \max_\theta \ell(\mathcal{D} : G, \theta) - \frac{\log M}{2} \text{Dim}[G]$$

where $M$ is the number of instances in $\mathcal{D}$, and $\text{Dim}[G]$ is the number of parameters in $G$. The BIC score is actually an approximation to the more principled full Bayesian score, that integrates over all possible parameterizations of the CPDs. A closed form for the Bayesian score, with a suitable prior, is known for Gaussian networks [9], but not for non-linear CPDs. Since our motivation comes mainly from improving non-linear network structure search, we will focus on the BIC score in the next sections.

A common search procedure for optimizing the score is the greedy hill-climbing. This procedure can be augmented with mechanisms for escaping local maxima, such as random walk perturbations upon reaching a local maxima and using a TABU list.

## 3　The "Ideal parent" Concept

Structure learning typically involves a traversal of a superexponential search space. Even when using a greedy local search procedure, structure learning is extremely time consuming. This problem is even more acute when considering parameterizations that require using non-linear optimization for estimating the maximum likelihood parameters. Our goal in this work is to efficiently select the approximately best candidate parent during the search procedure for Bayesian networks with continuous variables, and to facilitate the addition of new and effective hidden variables into the network structure. We start by characterizing the notion of an ideal parent that will enable us to do so.

### 3.1　Basic Framework

The basic idea is straightforward — for a given variable, we want to construct a hypothetical "ideal parent" that would best predict the variable. We will then compare existing candidate parents to this imaginary one and find the ones that are most similar. Throughout this discussion, we focus on a single variable $X$ that in the current network has parents $\mathbf{U}$. To make our discussion concrete, we focus on networks where we represent $X$ as a function of its parents $\mathbf{U} = \{U_1, \ldots, U_k\}$ with CPDs that have the following general form:

$$X = g(\alpha_1 \mathbf{u}_1, \ldots, \alpha_k \mathbf{u}_k : \theta) + \epsilon \quad (1)$$

where $g$ is a function that integrates the contributions of the parents with additional parameters $\theta$, $\alpha_i$ that are scale parameters applied to each of the parents, and $\epsilon$ that is a noise random variable with zero mean. In here, we assume that $\epsilon$ is Gaussian with variance $\sigma^2$.

When the function $g$ is the sum of its arguments, this CPD is the standard linear Gaussian CPD. However, we can also consider non-linear choices of $g$. For example,

$$g(y_1, \ldots, y_k : \theta) \equiv \theta_1 \frac{1}{1 + e^{-\sum_i y_i}} + \theta_0 \qquad (2)$$

is a sigmoid function where the response of $X$ to its parents' values is saturated when the sum is far from $0$. Both of these examples of CPDs are instances of *generalized linear models* (GLMs) [14]. This class of CPDs uses a function $g$ that is applied to the sum of its arguments, called the *link function* in the GLM literature. However, we can also consider more complex functions, as long as they are well defined for any desired number of parents. For example, in [16] models based on chemical reactions are considered, where the function $g$ does not have a GLM form. An example of a two variable function of this type is:

$$g(y_1, y_2 : \theta) = \theta \frac{y_1 y_2}{(1 + y_1)(1 + y_2)}$$

We also note that GLM literature deals extensively with different forms of noise. While we focus here on the case of additive Gaussian noise, the ideas we propose here can be extended to many of these noise distributions.

With this definition for a CPD, we now can define the ideal parent for $X$.

**Definition 3.1:** Given a dataset $\mathcal{D}$, and a CPD $g$ for $X$ given its parents $\mathbf{U}$ with parameters $\theta$ and $\alpha$, the *ideal parent* $Y$ of $X$ is such that for each instance $m$,

$$x[m] = g(\alpha_1 u_1[m], \ldots, \alpha_k u_k[m], y[m] : \theta) \qquad (3)$$

∎

Under mild conditions, the *ideal parent profile* (i.e., value of $Y$ in each instance) can be computed for almost any unimodal parametric conditional distribution. The only requirement from $g$ is that it should be invertible w.r.t. each one of the parents.[1]

The resulting profile for the ideal parent $Y$ is the optimal set of values for the $k + 1$'th parent, in the sense that it would maximize the likelihood of the child variable $X$. Intuitively, if we can efficiently find a candidate parent that is similar to the optimal parent, we can improve the model by adding an edge from this parent to $X$.

We now develop the computation of the similarity measure between the ideal profile and a candidate parent for the case of a linear Gaussian distribution. In Section 4 we describe how to use this measure to perform the actual search.

---

[1] In Definition 3.1 we implicitly assume $x[m]$ lies in the image of $g$. If this is not the case, we can substitute $x[m]$ with $x_g[m]$, the point in $g$'s image closest to $x[m]$. This guarantees the prediction's mode for the current set of parents and parameters is as close as possible to $X$.

### 3.2 Linear Gaussian

Let $X$ be a variable in the network with a set of parents $\mathbf{U}$, and a *linear Gaussian* conditional distribution. In this case, $g$ in Eq. (1) takes the form

$$g(\alpha_1 \mathbf{u}_1, \ldots, \alpha_k \mathbf{u}_k : \theta) \equiv \sum_i \alpha_i \mathbf{u}_i + \theta_0$$

In using the BIC score (see Section 2), whenever we consider a change in the structure, such as adding $Z$ as a new parent of $X$ whose current parents are $\mathbf{U}$, we need to compute the change in likelihood

$$\Delta_{X|\mathbf{U}}(Z) = \max_{\theta'} \ell_X(\mathcal{D} : \mathbf{U} \cup \{Z\}, \theta') - \ell_X(\mathcal{D} : \mathbf{U}, \theta)$$

where $\theta$ is the current maximum likelihood parameters for the family. The change in the BIC score is this difference combined with the change in penalty terms. To evaluate this difference, we need to compute the maximum likelihood parameters of $X$ given the new choice of parents. In the "ideal parent" approach we use a fast method for choosing the promising candidates for additional parent, and then compute the BIC score only for these candidates. This reduces the time complexity of the scoring step.

To choose promising candidate parents to add, we start by computing the ideal parent $Y$ for $X$ given its current set of parents. This is done by inverting the linear link function $g$ with respect to this additional parent $Y$ (note that we can assume, without loss of generality, that the scale parameter of $Y$ is 1). This results in

$$y[m] = x[m] - \sum_j \alpha_j u_j[m] - \theta_0 \qquad (4)$$

We can summarize this in vector notation, by using $\vec{x} = \langle x[1], \ldots, x[M] \rangle$, and so we get

$$\vec{y} = \vec{x} - \mathcal{U}\vec{\alpha}$$

where $\mathcal{U}$ is the matrix of parent values on all instances, and $\vec{\alpha}$ is the vector of scale parameters.

Having computed the *ideal parent profile*, we now want to efficiently evaluate its similarity to profiles of candidate parents. Intuitively, we want the similarity measure to reflect the likelihood gain by adding $P$ as a parent of $X$. Ideally, we want to evaluate $\Delta_{X|\mathbf{U}}(Z)$ for each candidate parent $Z$. However, instead of reestimating all the parameters of the CPD after adding $Z$ as a parent, we approximate this difference by only fitting the scaling factor associated with the new parent and freezing all other parameters of the CPD.

**Proposition 3.2** *Suppose that $X$ has parents $\mathbf{U}$ with a set $\vec{\alpha}$ of scaling factors. Let $Y$ be the ideal parent as described above, and $Z$ be some candidate parent. Then the change in likelihood of $X$ in the data, when adding $Z$ as a parent of*

$X$, while freezing all parameters except the scaling factor of $Z$, is

$$\begin{aligned} C_1(\vec{y}, \vec{z}) &\equiv \max_{\alpha_Z} \ell_X(\mathcal{D} : \mathbf{U} \cup \{Z\}, \theta \cup \{\alpha_Z\}) \\ &\quad - \ell_X(\mathcal{D} : \mathbf{U}, \theta) \\ &= \frac{1}{2\sigma^2} \frac{(\vec{y} \cdot \vec{z})^2}{\vec{z} \cdot \vec{z}} \end{aligned}$$

where the result follows from replacing $\alpha_Z$ with its *maximum likelihood* estimate which is $\frac{\vec{z} \cdot \vec{y}}{\vec{z} \cdot \vec{z}}$.

Note that by definition of $y$, the maximum likelihood estimator of $\sigma^2$ is $\frac{1}{M} \vec{y} \cdot \vec{y}$.[2] Thus, there is an intuitive geometric interpretation to the measure $C(\vec{y}, \vec{z})$ — it is proportional to the angle between $\vec{y}$ and $\vec{z}$. And so, we prefer a profile $\vec{z}$ that is similar to the ideal parent profile $\vec{y}$, regardless of its norm: It can easily be shown that $\vec{z} = c\vec{y}$ (for any constant $c$) maximizes this similarity measure.

Note that $C_1(\vec{y}, \vec{z})$ is a *lower bound* on $\Delta_{X|\mathbf{U}}(Z)$, the improvement on the log-likelihood by adding $Z$ as a parent of $X$. When we add the parent we optimize all the parameters, and so we expect to attain a likelihood as high, or higher, than the one we attain by freezing some of the parameters. This is illustrated in Fig. 1(a) that plots $C_1$ vs. the true likelihood improvement for several thousand edge modifications.

We can get a better lower bound by optimizing additional parameters. In particular, after adding a new parent, the errors in predictions change, and so we can readjust the variance term. As it turns out, we can perform this readjustment in closed form.

**Proposition 3.3** *Suppose that $X$ has parents $\mathbf{U}$ with a set $\vec{\alpha}$ of scaling factors. Let $Y$ be the ideal parent as described above, and $Z$ be some candidate parent. Then the change in likelihood of $X$ in the data, when adding $Z$ as a parent of $X$, while freezing all other parameters except the variance of $X$, is*

$$\begin{aligned} C_2(\vec{y}, \vec{z}) &\equiv \max_{\alpha_Z, \sigma} \ell_X(\mathcal{D} : \mathbf{U} \cup \{Z\}, \theta \cup \{\alpha_Z\}) \\ &\quad - \ell_X(\mathcal{D} : \mathbf{U}, \theta) \\ &= \frac{M}{2} \log \frac{1}{1 - \frac{(\vec{z} \cdot \vec{y})^2}{(\vec{z} \cdot \vec{z})(\vec{y} \cdot \vec{y})}} \\ &= -\frac{M}{2} \log \sin^2 \phi_{\vec{y}, \vec{z}} \end{aligned}$$

*where $\phi_{\vec{y}, \vec{z}}$ is the angle between $\vec{y}$ and $\vec{z}$.*

We can easily show that

$$C_1(\vec{y}, \vec{z}) \leq C_2(\vec{y}, \vec{z}) \leq \Delta_{X|\mathbf{U}}(Z)$$

due to the set of parameters we freeze in the optimization of each quantity. It is important to note that both $C_1$ and $C_2$

---

[2] We choose to use $\sigma^2$ explicitly in the definition of $C_1$ for compatibility with the development below.

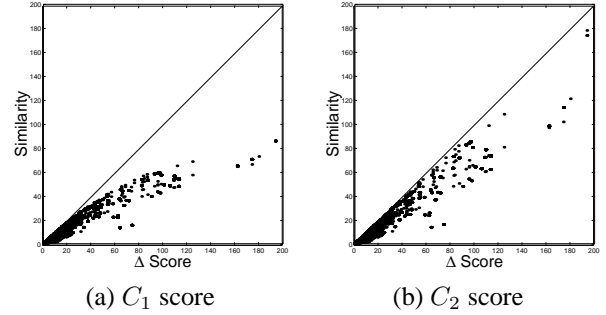

(a) $C_1$ score  (b) $C_2$ score

Figure 1: Demonstration of the $C_1$ (a) and $C_2$ (b) similarity measures for linear Gaussian CPDs. The similarity measures ($y$-axis) are shown against the change in log likelihood resulting from the corresponding edge modifications ($x$-axis). Points shown correspond to several thousand edge modifications in a run of the ideal parent method on real-life yeast gene expressions data.

are monotonic functions of $\frac{(\vec{y} \cdot \vec{z})^2}{\vec{z} \cdot \vec{z}}$, and so they consistently rank candidate parents of the same variable. As seen in Fig. 1, $C_2$ is clearly a tighter bound than $C_1$, particularly for promising candidates. When we compare changes that involve different ideal parents, such as adding a parent to $X_1$ compared to adding a parent to $X_2$, the ranking by these two measures might differ.

## 4 Ideal Parents in Search

The technical developments of the previous section show that we can approximate the score of candidate parents for $X$ by comparing them to the ideal parent $Y$. Is this approximate evaluation useful? We now discuss two ways of using this approximation during structure search.

### 4.1 Guiding Heuristic Search

When performing a local heuristic search, at each iteration we have a current candidate structure and we consider some operations on that structure. These operations might include edge addition, edge replacement, edge reversal and edge deletion. We can readily use the ideal profiles and similarity measures developed to speed up two of these: edge addition and edge replacement. These two modification form the bulk of edge changes considered by a typical search algorithm, since edge deletions and reversals can only be applied to the relatively small number of existing edges.

When considering adding an edge $Z \rightarrow X$, we use the ideal parent profile for $X$ and compute its similarity to $Z$. We repeat this for every other candidate parent for $X$. We then compute the full score only for the $K$ most similar candidates, and insert them (and the associated change in score) to a queue of potential operations. In a similar way, we can utilize the ideal parent profile for considering edge replacement for $X$. Suppose that $Z$ is a parent of $X$. We can define the ideal profile for replacing $Z$ while freezing all other parameters of the CPD of $X$. The difference here is that for each current parent of $X$ we compute a separate ideal profile - one corresponding to replacement of that parent with a new one. We then use the same policy as above for examining replacement of each one of the parents.

We note that we can tradeoff between the accuracy of our move evaluations and the speed of the search, by changing $K$, the number of candidate changes per family for which we compute a full score. Using $K = 1$, we only score the best candidate according to the ideal parent method ranking, thus achieving the largest speedup. Since our ranking only approximates the true score difference, this strategy might miss good moves. Using higher values of $K$ brings us closer to the standard search both in terms of move selection quality but also in terms of computation time.

In the experiments in Section 6, we integrated the changes described above into a greedy hill climbing heuristic search procedure. This procedure also examines moves that remove an edge, which we evaluate in the standard way. The greedy hill climbing procedure applies the best available move at each iteration (among these that were chosen for full evaluation).

### 4.2 Adding New Hidden Variables

One of the hardest challenges in learning graphical models is dealing with *hidden* variables. Such variables pose several problems, the hardest of which is detecting when and how should one add a new hidden variable into the network structure. When we are learning networks in a domain with a large number of variables, and each hidden variable influences a relatively small subset of these variables, this becomes a major issue (see *e.g.*, [4, 3, 13, 21]).

The ideal parent profiles provide a straightforward way to find when and where to add hidden variables to the domain. The intuition is fairly simple: if the ideal parents of several variables are similar to each other, then we know that a similar input is predictive of all of them. Moreover, if we do not find a variable in the network that is close to these ideal parents, then we can consider adding a new hidden variable that will serve as their combined input, and, in addition, have an informed initial estimate of its profile.

To introduce a new hidden variable, we would like to require that it will be beneficial for several children at once. The difference in log-likelihood due to adding a new parent with profile $\vec{z}$ is the difference between the log-likelihood of families it is involved in:

$$\Delta_{X_1,\ldots,X_L}(Z) = \sum_{i}^{L} \Delta_{X_i|\mathbf{U}_i}(Z)$$

where we assume, without loss of generality, that the members of the cluster are $X_1, \ldots, X_L$. To score the network with $Z$ as a new hidden variable, we also we need to deal with the difference in the complexity penalty term, and the likelihood of $Z$ variable as a root variable. These terms, however, can be readily evaluated. The difficult term is finding the profile $\vec{z}$ that maximizes $\Delta_{X_1,\ldots,X_L}(Z)$.

Using the ideal parent approximation, we can lower bound this improvement by

$$\sum_{i}^{L} C_1(\vec{y}_i, \vec{z}) \equiv \sum_{i} \frac{1}{2\sigma_i^2} \frac{(\vec{z} \cdot \vec{y}_i)^2}{\vec{z} \cdot \vec{z}} \quad (5)$$

and so we want to find $\vec{z^*}$ that maximizes this bound. We will then use this optimized bound as our cluster score. That is we want to find $\vec{z^*}$ that maximizes

$$\vec{z^*} = \arg\max_{\vec{z}} \frac{\vec{z}^T \mathcal{Y}\mathcal{Y}^T \vec{z}}{\vec{z}^T \vec{z}} \quad (6)$$

where $\mathcal{Y}$ is the matrix whose columns are $y_i/\sigma_i$.

It is easy to see that $\vec{z^*}$ must lie in the span of $\mathcal{Y}$: any component orthogonal to this span increases the denominator of Eq. (6), but leaves the numerator unchanged, and therefore does not obtain a maximum. We can therefore express the solution as:

$$\vec{z^*} = \sum_{i} \lambda_i y_i / \sigma_i = \mathcal{Y}\vec{\lambda} \quad (7)$$

Furthermore, the objective in Eq. (6) is known as the *Rayleigh quotient* of the matrix $\mathcal{Y}\mathcal{Y}^T$ and the vector $\vec{z}$, and its optimum is achieved when $\vec{z}$ equals the eigenvector of $\mathcal{Y}\mathcal{Y}^T$ corresponding to the largest eigenvalue [20]. Using Eq. (7) we can express this eigenvector problem as follows:

$$\mathcal{Y}\mathcal{Y}^T \vec{z^*} = \gamma \vec{z^*}$$

Plugging in Eq. (7) and defining $A = \mathcal{Y}^T \mathcal{Y}$, we write

$$AA\vec{\lambda} = \gamma A\vec{\lambda}$$

We can now either solve this reduced generalized eigenvalue problem directly, or, if $A$ is non-singular, we can multiply both sides by $A^{-1}$ and end up with a simple eigenvalue problem:

$$A\vec{\lambda} = \gamma \vec{\lambda}$$

which is easy to solve as the dimension of $A$ is $L$, the number of variables in the cluster, which is typically relatively small. Once we find the $L$ dimensional eigenvector $\lambda^*$ with the largest eigenvalue $\gamma^*$, we can express with it the desired parent profile $\vec{z^*}$.

We can get a better bound of $\Delta_{X_1,\ldots,X_L}(Z)$ if we use $C_2$ similarity rather than $C_1$. Unfortunately, optimizing the profile $\vec{z}$ with respect to this similarity measure is a harder problem that is not solvable in closed form. Since the goal of the cluster identification is to provide a good starting point for the following iterations that will eventually adapt the structure, we use the closed form solution for Eq. (6). Note that once we optimized the profile $z$ using the above derivation, we can still use the $C_2$ similarity score to provide a better bound on the quality of this profile as a new parent for $X_1, \ldots, X_L$.

Now that we can approximate the benefit of adding a hidden variable to a cluster of variables, we still need to consider different clusters to find the most beneficial one. As the number of clusters is exponential, we adapt a heuristic *agglomerative clustering* approach (*e.g.*, [2]) to explore different clusters. We start with each ideal parent profile as an individual cluster and at each point we merge the two clusters that lead to the best expected improvement in the BIC score (combining the above approximation with the change in penalty terms). This procedure potentially involves $O(N^3)$ merges, where $N$ is the number of variables. We save much of the computations by pre-computing the matrix $\mathcal{Y}^T\mathcal{Y}$ only once, and then using the relevant submatrix in each merge In practice, the time spent in this step is insignificant in the overall search procedure.

### 4.3 Learning with Missing Values

Once we consider learning structure with hidden variables, we have to deal with the issue of missing values while considering subsequent structure changes. Similar considerations can arise if the dataset contains partial observations of some of the variables.

To deal with this problem, we use an Expectation Maximization approach [1] and its application to network structure learning [5]. At each step in the search we have a current network that provides an estimate of the distribution that generated the data, and use it to compute a distribution over possible completions of the data. Instead of maximizing the BIC score, we attempt to maximize the expected BIC score

$$\boldsymbol{E}_Q[BIC(\mathcal{D}, G) \mid \mathcal{D}_o] = \sum Q(\mathcal{D} \mid \mathcal{D}_o) BIC(\mathcal{D}, G)$$

where $\mathcal{D}_o$ is the observed data, and $Q$ is the distribution represented by the current network. As the BIC score is a sum over local terms, we can use linearity of expectations to rewrite this objective as a sum of expectations, each over the scope of a single CPD. This implies that when learning with missing values, we need to use the current network to compute the posterior distribution over the values of variables in each CPD we consider. Using these posterior distributions we can estimate the expectation of each local score, and use them in standard structure search (discussed above). Once the search algorithm converges, we use the new network for computing expectations and reiterate until convergence (see [5]).

How can we combine the ideal parent method into this structural EM search? Since we do not necessarily observe neither $X$ nor all of its parents, the definition of ideal parent cannot be applied directly. Instead, we define the ideal parent to be the profile that will match the expectations given $Q$. That is, we choose $y[m]$ so that

$$\boldsymbol{E}_Q[x[m] \mid \mathcal{D}_o] = \\ \boldsymbol{E}_Q[g(\alpha_1 u_1[m], \ldots, \alpha_k u_k[m], y[m] : \theta) \mid \mathcal{D}_o]$$

In the case of linear CPDs, this implies that

$$\vec{y} = \boldsymbol{E}_Q[\vec{x} \mid \mathcal{D}_o] - \boldsymbol{E}_Q[\mathcal{U} \mid \mathcal{D}_o]\vec{\alpha}$$

Once we define the ideal parent, we can use it to approximate changes in the expected BIC score (given $Q$). For the case of linear Gaussian, we get terms that are similar to $C_1$ and $C_2$ of Proposition 3.2 and Proposition 3.3, respectively. The only change is that we apply the similarity measure on the expected value of $\vec{z}$ for each candidate parent $Z$. This is in contrast to exact evaluation of $\boldsymbol{E}_Q\left[\Delta_{X|\mathbf{U}}Z \mid \mathcal{D}_o\right]$, which requires the computation of the expected sufficient statistics of $\mathbf{U}$, $X$, and $Z$.

To facilitate efficient computation, we adopt an approximate variational *mean-field* form (*e.g.*, [10, 15]) for the posterior. This approximation is used both for the ideal parent method and the standard greedy approach used in Section 6. This results in computations that require only the first and second moments for each instance $z[m]$, and thus can be easily obtained from $Q$.

Finally, we note the structural EM iterations are still guaranteed to converge to a local maximum. In fact, this does *not* depend on the fact that $C_1$ and $C_2$ are lower bounds of the true change to the score, since these measures are only used to pre-select promising candidates which are scored before actually being considered by the search algorithm. Indeed, the ideal parent method is a modular structure candidate selection algorithm and can be used as a black-box by any search algorithm.

## 5 Non-linear CPDs

We now turn to dealing with the case of non-linear CPDs. In the class of CPDs we are considering, this non-linearity is mediated by the function $g$, which we assume here to be invertible. Examples of such function include the sigmoid function shown in Eq. (2) and hyperbolic functions that are suitable for modeling gene transcription regulation [16], among many others. When we learn with non-linear CPDs, parameter estimation is harder. To evaluate a potential parent $P$ for $X$ we have to perform non-linear optimization (*e.g.*, conjugate gradient) of all of the $\alpha$ coefficients of all parents as well as other parameters of $g$. In this case, a fast approximation can boost the computational cost of the search significantly.

As in the case of linear CPDs, we compute the ideal parent profile $\vec{y}$ by inverting $g$ (We assume that the inversion of $g$ can be performed in time that is proportional to the calculation of $g$ itself as is the case in CPDs considered above.) Suppose we are considering the addition of a parent to $X$ in addition to its current parents $\mathbf{U}$, and that we have computed the value of the ideal parent $y[m]$ for each sample $m$ by inversion of $g$. Now consider a particular candidate parent $Z$ whose value at the $m$'th instance is $Z[m]$. How will the difference between the ideal value and the value of $Z$ reflect in prediction of $X$ for this instance?

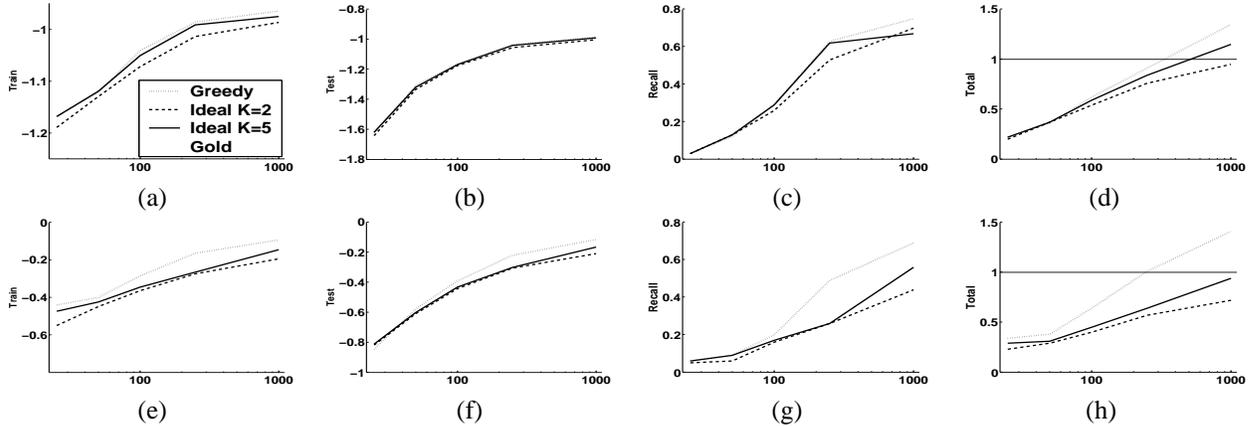

Figure 2: Evaluation of using **Ideal** search on synthetic data with 44 variables. We compare **Ideal** search with $K = 2$ (dashed) and $K = 5$ (solid), against the standard **Greedy** procedure (dotted). The figures show, as a function of the number of instances ($x$-axis), for linear Gaussian CPDs: (a) average training log likelihood per instance per variable; (b) same for test; (c) percent of true edges obtained in learned structure; (d) total number of edges learned as percent of true number. (e)-(h) repeat the above for Sigmoid CPDs.

In the linear case, the difference $z[m] - y[m]$ translated through $g$ to a prediction error. In the non-linear case, the effect of difference on predicting $X$ depends on other factors, such as the values of the other parents, despite the fact that their parameters are held fixed: Consider again the sigmoid function $g$ of Eq. (2). If the sum of the arguments to $g$ are close to 0, then $g$ locally behaves like a sum of its arguments. On the other hand, if the sum is far from 0, the function is in one of the saturated regions, and big differences in the input almost do not change the prediction.

Our solution to this problem is to approximate $g$ by a linear function around the value we predict. We use a first-order Taylor expansion of $g$ around the value of $\vec{y}$ and write

$$g(\mathbf{u}, \vec{z}) \approx g(\mathbf{u}, \vec{y}) + (\vec{z} - \vec{y}) \frac{\partial g(\mathbf{u}, \vec{y})}{\partial \vec{y}}$$

As a result, the "penalty" for a distance between $\vec{z}$ and $\vec{y}$ depends on the gradient of $g$ at the particular value of $\vec{y}$, given the value of the other parents. In instances where the derivative is small, larger deviations between $y$ and $z$ have little impact on the likelihood of $X$, and in other instances where the derivative is large, the same deviations may lead to worse likelihood.

With this linear approximation, we can develop similarity measures that parallel the developments for the linear case. There are two main differences. First, due to the use of Taylor expansion, we can no longer prove that these are underestimates of the likelihood. Second, due to the different value of the gradient at different instances, the contribution of distance at each instance will be weighted differently.

**Proposition 5.1** *Suppose that $X$ has parents* $\mathbf{U}$ *with a set $\vec{\alpha}$ of scaling factors. Let $Y$ be the ideal parent as described above, and $Z$ be some candidate parent. Then the change in log-likelihood of $X$ in the data, when adding $P$ as a parent of $X$, while freezing all other parameters, is approximately*

$$C_1(\vec{y} \star g'(y), \vec{z} \star g'(y)) - \frac{1}{2\sigma^2}(c_1 - c_2). \quad (8)$$

*where $g'(y)$ is the vector whose $m$'th component is $\partial g(\vec{\alpha}\mathbf{u}, y)/\partial y \mid_{\mathbf{u}[m], y[m]}$, and $\star$ denotes component-wise product. Similarly, if we also optimize the variance, then the change in log-likelihood is approximately*

$$C_2(\vec{y} \star g'(y), \vec{z} \star g'(y)) - \frac{M}{2} \log \frac{c_1}{c_2} \quad (9)$$

*In both cases,*

$$c_1 = (\vec{y} \star g'(y)) \cdot (\vec{y} \star g'(y)) \, ; \, c_2 = (\vec{x} - g(\mathbf{u})) \cdot (\vec{x} - g(\mathbf{u}))$$

*do not depend on $\vec{z}$.*

Thus, we can use exactly the same measures as before, except that we "distort" the geometry with the weight vector $g'(y)$ that determines the importance of different instances. To approximate the likelihood difference, we also add the correction term which is a function of $c_1$ and $c_2$. This correction is not necessary when comparing two candidates for the same family, but is required for comparing candidates from different families, or when adding hidden values. Note that if $g$ is linear then the correction term vanishes altogether and Proposition 3.2 and Proposition 3.3 are recovered.

We can now efficiently perform the ideal *add* and *replace* steps in the structure search. The complexity is again $O(M(d-1))$ for computing the ideal profile and $O(KM)$ for computing the similarity measure for each candidate parent. As before, the significant gain in speed is that we only perform few parameter optimizations (that are expected to be costly as the number of parents grows), rather than $O(N)$ such optimizations.

| Dataset | vars | instances | Ideal $K = 2$ vs Greedy | | | | | | Ideal $K = 5$ vs Greedy | | | | | |
|---|---|---|---|---|---|---|---|---|---|---|---|---|---|---|
| | | | train | test | edges | moves | eval | speedup | train | test | edges | moves | eval | speedup |
| Linear Gaussian | | | | | | | | | | | | | | |
| AA nomiss | 44 | 173 | -0.024 | 0.006 | 87.1 | 96.5 | 3.6 | 2 | -0.008 | 0.007 | 94.9 | 96.5 | 9.3 | 2 |
| AA nomiss Cond | 173 | 44 | -0.038 | 0.082 | 92.2 | 92.6 | 1.2 | 2 | -0.009 | 0.029 | 96.9 | 98.2 | 2.9 | 2 |
| Met nomiss | 89 | 173 | -0.033 | -0.024 | 88.7 | 91.5 | 1.6 | 3 | -0.013 | -0.016 | 94.5 | 96.9 | 4.4 | 2 |
| Met nomiss Cond | 173 | 89 | -0.035 | -0.015 | 91.3 | 98.0 | 1.0 | 2 | -0.007 | -0.023 | 98.9 | 98.5 | 2.4 | 2 |
| Linear Gaussian with missing values | | | | | | | | | | | | | | |
| AA | 354 | 173 | -0.101 | -0.034 | 81.3 | 95.2 | 0.4 | 5 | -0.048 | -0.022 | 90.7 | 96.0 | 0.9 | 5 |
| AA Cond | 173 | 354 | -0.066 | -0.037 | 74.7 | 87.5 | 0.4 | 14 | -0.033 | -0.021 | 86.3 | 101.1 | 1.6 | 11 |
| Sigmoid | | | | | | | | | | | | | | |
| AA nomiss | 44 | 173 | -0.132 | -0.065 | 49.7 | 59.4 | 2.0 | 38 | -0.103 | -0.046 | 60.4 | 77.6 | 6.1 | 18 |
| AA nomiss Cond | 173 | 44 | -0.218 | 0.122 | 62.3 | 76.7 | 1.0 | 36 | -0.150 | 0.103 | 73.7 | 79.4 | 2.3 | 21 |
| Met nomiss | 89 | 173 | -0.192 | -0.084 | 47.9 | 58.3 | 0.9 | 65 | -0.158 | -0.059 | 56.6 | 69.8 | 2.6 | 29 |
| Met nomiss Cond | 173 | 89 | -0.207 | -0.030 | 60.5 | 69.5 | 0.8 | 53 | -0.156 | -0.042 | 69.8 | 77.7 | 2.2 | 29 |

Table 1: Performance comparison of the Ideal parent search with $K = 2$ or $K = 5$ and Greedy on real data sets. *vars* - number of variables in the dataset; *train* - average difference in training set log likelihood per instance per variable; *test* - same for test set; *edges* - percent of edges learned by Ideal with respect to those learned by Greedy. *moves* - percent of structure modifications taken during the search; *eval* - percent of moves evaluated; *speedup* - speedup of Ideal over greedy method. All figures are average over 5 fold cross validation sets.

Adding a new hidden variable with non-linear CPDs introduces further complication. We want to use, similarly to the case of a linear model, the structure score of Eq. (5) with the distorted $C_1$ measure. Optimizing this measure has no closed form solution in this case and we need to resort to an iterative procedure or an alternative approximation. In here, we approximate Eq. (8) with a form that is similar to the linear Gaussian case, with the "distorted" geometry of $\vec{y}$.

## 6 Experiments

We now examine the impact of the ideal parent method in two settings. In the first setting, we use this method for pruning the number of potential moves that are evaluated by greedy hill climbing structure search. We apply this learning procedure to complete data (and data with some missing values) to learn dependencies between the observed variables. In the second setting, we use the ideal parent method as a way of introducing new hidden variables, and also use it as a guide to reduce the number of evaluations when learning structure that involves hidden variables and observed ones with a Structural EM search procedure.

In the first setting, we applied standard greedy hill climbing search (Greedy) and greedy hill climbing supplemented by the ideal parent method as discussed in Section 4 (Ideal). In using the ideal parent method, we used the $C_2$ similarity measure (Section 3) to rank candidate edge additions and replacements, and then applied full scoring only to the top $K$ ranking candidates per variable.

To evaluate the impact of the method, we start with a synthetic experiment where we know the true underlying network structure. In this setting we can evaluate the magnitude of the performance cost resulting from the approximation we use. We used a network learned from real data (see below) with 44 variables. From this network we can generate datasets of different sizes and apply our method with different values of $K$. Fig. 2 compares the ideal parent method and the standard greedy procedure for linear Gaussian CPDs (upper panel) and Sigmoid CPDs (lower panel). Using $K = 5$ is, as we expect, closer to the performance of the standard greedy method both in terms of training set Fig. 2(a,e) and test set Fig. 2(b,f) performance then $K = 2$. For linear Gaussian CPDs test performance is essentially the same with a slight advantage for the standard greedy method using Sigmoid CPDs. When considering the percent of true edges recovered Fig. 2(c,g), again the standard method shows some advantage over the ideal method with $K = 5$. However, looking at the total number of edges learned Fig. 2(d,h), we can see that the standard greedy method achieves this by using close to 50% more edges then the original structure for Sigmoid CPDs. Thus, advantage in performance comes at a high complexity price (and as we demonstrate below, at a significant speed cost).

We now examine the effect of the method on learning from real-life datasets. We base our datasets on a study that measures the expression of the baker's yeast genes in 173 experiments [8]. In this study, researchers measured expression of 6152 yeast genes in its response to changes in the environmental conditions, resulting in a matrix of $173 \times 6152$ measurements. In the following, for practical reasons, we use two sets of genes. The first set consists of 639 genes that participate in general metabolic processes (Met), and the second is a subset of the first with 354 genes which are specific to amino acid metabolism (AA). We choose these sets since part of the response of the yeast to changes in its environment is in altering the activity levels of different parts of its metabolism. For some of the experiments below, we focused on subsets of genes for which there are no missing values (nomiss, consisting of 89 and 44 genes, respectively).

On these datasets we can consider two tasks. In the first, we treat genes as variables and experiments as instances. The learned networks indicate possible regulatory or functional connections between genes [6]. A complementary task is to treat experiments as variables (Cond). A learned network in this scenario indicates dependencies between the responses to different conditions.

In Table 1 we summarize differences between the Greedy search and the Ideal search with $K$ set to 2 and 5, for the linear Gaussian CPDs as well as sigmoid CPDs. Since the $C_2$ similarity is only a lower bound of the $BIC$ score difference, we expect the candidate ranking of the two to be different. As most of the difference comes from freezing some of the parameters, a possible outcome is that the Ideal search is less prone to over-fitting. Indeed as we see, though the training set log likelihood in most cases is lower for Ideal search, the test set performance is comparable or better.

Of particular interest is the tradeoff between accuracy and speed when using the ideal parent method. In Fig. 3 we examine this tradeoff in four of the data sets described above using linear Gaussian and sigmoid CPDs. In both cases, the performance of the ideal parent method approaches that of standard greedy as $K$ is increased. As we can expect, in both types of CPDs the ideal parent method is faster even for $K = 5$. However, the effect on total run time is much more pronounced when learning networks with non-linear CPDs. In this case, most of the computation is spent in optimizing the parameters for scoring candidates. And so, reducing the number of candidates evaluated results in a dramatic effect. As Table 1 shows, the number of score evaluations with the ideal heuristic is usually a small fraction of the number of evaluations carried out by the standard search. This speedup in non-linear networks makes previously "intractable" real-life learning problems (like gene regulation network inference) more accessible.

In the second experimental setting, we examine the ability of our algorithm to learn structures that involve hidden variables and introduce new ones during the search. In this setting, we focus on *two layered networks* where the first layer consists of hidden variables, all of which are assumed to be roots, and the second layer consists of observed variables. Each of the observed variables is a leaf and can depend on one or more hidden variables. Learning such networks involves introducing different hidden variables, and determining for each observed variable which hidden variables it depends on.

To test the performance of our algorithm, we used a network topology that is curated [16] from biological literature for the regulation of cell-cycle genes in yeast. This network involves 7 hidden variables and 141 observed variables. We learned the parameters for the network (using either linear Gaussian CPDs or sigmoid CPDs) from a cell cycle gene expression dataset [19]. From the learned network we then sampled datasets of varying sizes, and tried to recreate the regulation structure using either greedy search or ideal parent search with the corresponding type of CPDs. In both search procedures we introduce hidden variables in a gradual manner. We start with a network where a sin-gle hidden variable is connected as the only parent to all observed variables. After parameter optimization, we introduce another hidden variable - either as a parent of all observed variables (in greedy search), or to members of the highest scoring cluster (in ideal parent search, as explained in Section 4.2). We then let the structure search modify edges (subject to the two-layer constraints) until no beneficial moves are found, at which point we introduce another hidden variable, and so on. The search terminates when it is no longer beneficial to add a new variable.

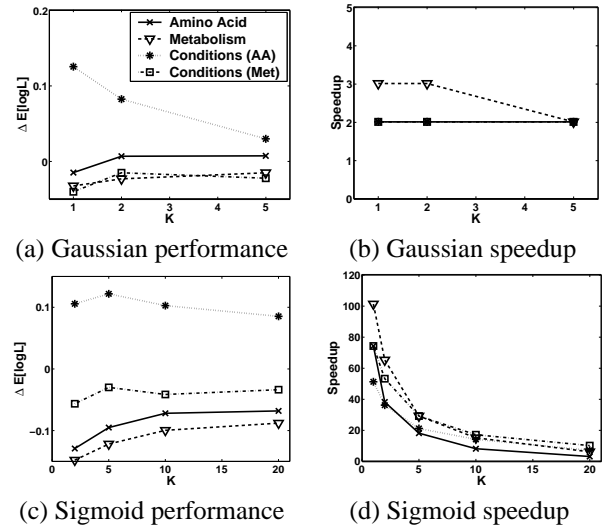

Figure 3: Evaluation of Ideal search on real-life data. (a) average log likelihood per instance on test data (based on 5-fold cross validataion) relative to Greedy when learning with linear Gaussian CPDs ($y$-axis) against $K$ ($x$-axis). (b) Speedup over Greedy ($y$-axis) against $K$ ($x$-axis). (c),(d) same for sigmoid CPDs.

Fig. 4 shows the performance of the ideal parent search and the standard greedy procedure as a function of the number of instances. As can be seen, although there are some differences in training set likelihood, the performance on test data is essentially similar, and approaches that of the Golden model (true structures with trained parameters) as the number of training instances grows. Thus, as in the case of the yeast experiments considered above, there was no degradation of performance due to the approximation made by our method.

We then considered the application of our algorithm to the real-life cell-cycle gene expression data described above with linear Gaussian CPDs. Although this data set contains only 17 samples, it is of high interest from a biological perspective to infer from it as much as possible on the structure of regulation. We performed leave-one-out cross validation and compared the Ideal parent method with $K = 2$ and $K = 5$ to the Greedy. To avoid over-fitting, we limited the number of hidden parents for each observed variable to 2. Although the standard greedy procedure achieved higher train log-likelihood performance, its test performance is significantly worse as a result of over-fitting for two par-ticular instances. As we have demonstrated in the synthetic example above, the ability of the ideal method to avoid over-fitting via a guided search does not come at the price of diminished performance when data is more plentiful. When the observed variables were allowed to have upto 5 parents, all methods demonstrated over-fitting which for Greedy was by far more severe.

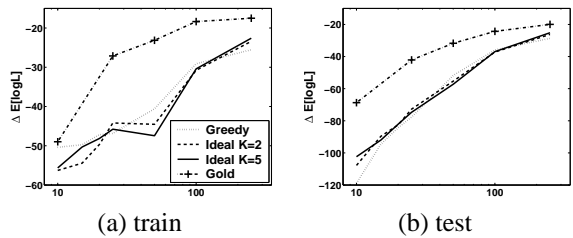

(a) train　　　　　　　　(b) test

Figure 4: Evaluation of performance in **two-layer network** experiments using synthetic data generated from a model learned from real data with 143 observable variables. (a) average log likelihood per instance on *training* data ($y$-axis) for Greedy , Ideal search with $K = 2$ and Ideal search with $K = 5$, and the true structure with trained parameters Golden, when learning with linear Gaussian CPDs against the number of training samples ($x$-axis). (b) Same for *test* set.

Finally, as we expect the nature of biological interactions in the cell cycle regulation domain to be non-linear, we set out to learn a model for this data using the richer but computationally demanding Sigmoid CPDs. To make computations tractable, we used in both methods a variational mean field approximation for computing posteriors over the hidden variables in the E-step of Structural EM. Unfortunately, even in this limited setting the standard greedy method proved non-feasible. The ideal method produced networks with interesting structure which, as shown in [16], can be subject to biological analysis.

## 7　Discussion and Future Work

In this work we set out to learn continuous variable networks. We addressed two fundamental challenges: First, we show how to speed up structure search, particularly for non-linear conditional probability distributions. This speedup is essential as it makes structure learning feasible in many interesting real life problems. Second, we show a principled way of introducing new hidden variables into the network structure. We use the concept of an "ideal parent" for both of these tasks and show its benefits on both synthetic and real-life biological domains.

The unique aspect of the Ideal parent approach is that it leverages on the parametric structure of the conditional distributions. In here, we applied this in conjunction with a greedy search algorithm. However, it can be supplemented to many other search procedures, such as simulated annealing, as a way of speeding up evaluation of candidate moves. Of particular interest is how our method can help algorithms that inherently limit the search to promising candidates such as the "Sparse Candidate" algorithm [7].

Few works touched on the issue of when and how to add a hidden variable in the network structure (*e.g.*, [4, 3, 13, 21]). Only some of these methods are potentially applicable to continuous variable networks, and none have been adapted to this context. To our knowledge, this is the first work to address this issue in a general context of continuous variable networks.

Many challenges remain. First, to combine the Ideal Parent method within other search procedures as a plug-in for candidate selection. Second, to apply the method to additional and more complex conditional probability distributions (e.g., [16]), and to leverage the connection to Generalized Linear Models [14], where a variety of optimization methods for specific types of CPDs exist. Finally, to use better approximations for adding new hidden variables in the non-linear case.

### Acknowledgements

We thank S. Shalev-Shwartz, A. Jaimovich, E. Portugaly, and the anonymous reviewers for comments on earlier versions of this manuscript. This work was supported, in part, by a grant from the Israeli Ministry of Science. I. Nachman and G. Elidan were also supported by the Horowitz fellowship. N. Friedman was also supported by the Harry & Abe Sherman Senior Lectureship in Computer Science and by the Bauer Center for Genomics Research, Harvard University.